\documentclass{llncs}
\usepackage{makeidx}  
\usepackage{amsmath,graphicx}
\usepackage[T1]{fontenc}
\usepackage[utf8]{inputenc}
\usepackage{subcaption}
\usepackage{booktabs}
\usepackage{color}
\usepackage{float}
\begin{document}
\frontmatter          
\pagestyle{headings}  
\addtocmark{CNN for detecting human sperm cells} 
\mainmatter              
\title{Convolutional neural networks for segmentation and object detection of human semen}
\titlerunning{CNN for detecting human sperm cells}  
%
\author{Malte S. Nissen\inst{1,2,3,4} \and Oswin Krause\inst{1} \and Kristian Almstrup\inst{2,3}
\and Søren Kjærulff\inst{4} \and Torben T. Nielsen\inst{4} \and Mads Nielsen\inst{1}}
\authorrunning{Nissen et al.} 
%
\tocauthor{Malte S. Nissen, Oswin Krause, Kristian Almstrup, Søren Kjærulff,
 Torben T. Nielsen, and Mads Nielsen}
\institute{Department of Computer Science, University of Copenhagen, Denmark \\ \email{nissen@di.ku.dk}
\and
Department of Growth and Reproduction, Rigshospitalet, University of Copenhagen, Denmark
\and
International Center for Research and Research Training in Endocrine Disruption of Male Reproduction and Child Health (EDMaRC), Rigshospitalet, University of Copenhagen, Denmark
\and
ChemoMetec A/S, Allerød, Denmark}

\maketitle              

\begin{abstract}
We compare a set of convolutional neural network (CNN) architectures for the
task of segmenting and detecting human sperm cells in an image taken from a semen sample.
In contrast to previous work, samples are not stained or washed to allow for full sperm 
quality analysis, making analysis harder due to clutter. Our results indicate that training on full images is superior to training on patches
when class-skew is properly handled. Full image training including up-sampling during training
proves to be beneficial in deep CNNs for pixel wise accuracy and detection performance.
Predicted sperm cells are found by using connected components on the CNN predictions.
We investigate optimization of a threshold parameter on the size of detected components.
Our best network achieves 93.87\% precision and 91.89\% recall on our test dataset after
thresholding outperforming a classical image analysis approach.

\end{abstract}
\section{Introduction}
\label{sec:introduction}
\begin{figure}[h!]
	\centering
	\begin{subfigure}[t]{0.19\columnwidth}
		\includegraphics[width=\textwidth]{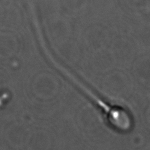}
		\caption{}
		\label{fig:in_focus}
	\end{subfigure}
	\begin{subfigure}[t]{0.19\columnwidth}
		\includegraphics[width=\textwidth]{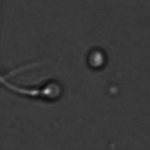}
		\caption{}
		\label{fig:sperm_and_debris}
	\end{subfigure}
	\begin{subfigure}[t]{0.19\columnwidth}
		\includegraphics[width=\textwidth]{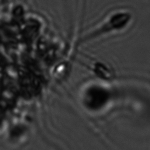}
		\caption{}
		\label{fig:aggregated}
	\end{subfigure}
	\begin{subfigure}[t]{0.19\columnwidth}
		\includegraphics[width=\textwidth]{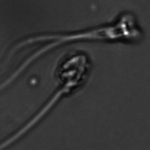}
		\caption{}
		\label{fig:agglutinated}
	\end{subfigure}
	\begin{subfigure}[t]{0.19\columnwidth}
		\includegraphics[width=\textwidth]{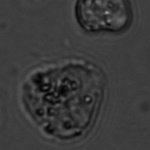}
		\caption{}
		\label{fig:round_cells}
	\end{subfigure}
	\begin{subfigure}[t]{0.19\columnwidth}
		\includegraphics[width=\textwidth]{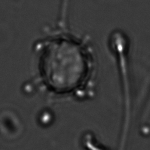}
		\caption{}
		\label{fig:round_cell_weird}
	\end{subfigure}
	\begin{subfigure}[t]{0.19\columnwidth}
		\includegraphics[width=\textwidth]{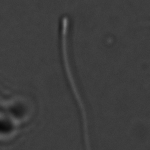}
		\caption{}
		\label{fig:headless}
	\end{subfigure}
	\begin{subfigure}[t]{0.19\columnwidth}
		\includegraphics[width=\textwidth]{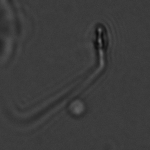}
		\caption{}
		\label{fig:deform}	
	\end{subfigure}
	\begin{subfigure}[t]{0.19\columnwidth}
		\includegraphics[width=\textwidth]{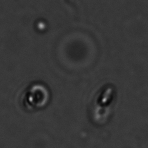}
		\caption{}
		\label{fig:deform_debris}
	\end{subfigure}
	\begin{subfigure}[t]{0.19\columnwidth}
		\includegraphics[width=\textwidth]{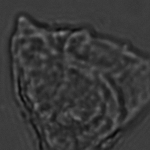}
		\caption{}
		\label{fig:debris}
	\end{subfigure}
	\caption{\label{fig:debris_all}Examples of debris, variations, and morphological abnormalities: normal sperm cell (\subref{fig:in_focus}, \subref{fig:sperm_and_debris}), aggregated cells out of focus (\subref{fig:aggregated}), agglutinated cells (\subref{fig:agglutinated}), round cells (\subref{fig:round_cells}, \subref{fig:round_cell_weird}), headless sperm (\subref{fig:headless}), sperm head seen from the side or morphologically abnormal (\subref{fig:deform}, \subref{fig:deform_debris}), circular tails (\subref{fig:deform_debris}), and other types of artifacts and debris (\subref{fig:sperm_and_debris}, \subref{fig:round_cell_weird}, \subref{fig:debris}).}
	\vspace{1em}
 	\includegraphics[width=0.98\columnwidth]{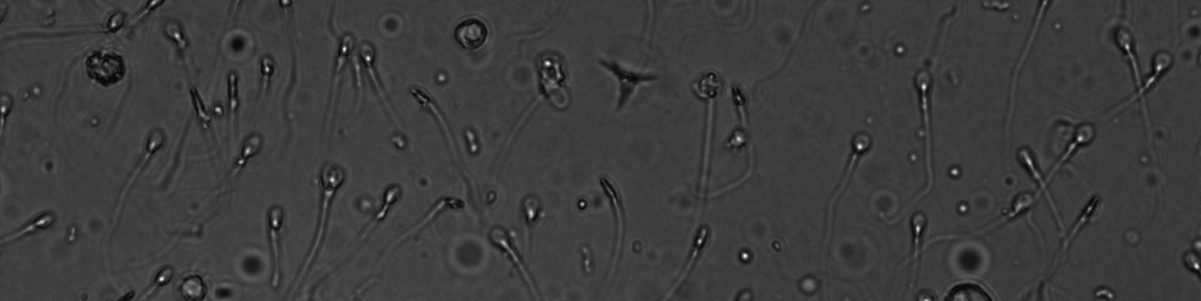}
 	\caption{$1200\times300$ pixel cut-out of image from the dataset}
 	\label{fig:example_big}
\end{figure}

Sperm Quality Analysis (SQA) involves measuring concentration, 
morphology, and motility \cite{world2010laboratory} of sperm cells. 
For the application to animal sperm cells, there exist a number of
commercial
Computer-Aided Sperm Analysis (CASA) systems, such as the 
Hamilton-Thorne \emph{IVOS-II} and 
\emph{CEROS-II}\footnote{\url{http://www.hamiltonthorne.com/}}
and the \emph{Sperm Class Analyzer}\footnote{\url{http://www.micropticsl.com/products/sperm-class-analyzer-casa-system/}}. 

Human semen samples have a significantly lower quality of sperm cells compared to most
animals \cite{van2015future},
which increases the accuracy demand on the analysis.
Moreover, human semen is often cluttered with debris and cells other than normal mature sperms. 
Fig. \ref{fig:debris_all} shows examples of typical debris, variations 
and morphological abnormalities of human sperm samples. Fig. \ref{fig:example_big} shows a section of a typical image.
\par In practice, staining and smearing are often used for preparation of samples
to highlight specific properties of the cells
\cite{bijar2012fully,carrillo2007computer,chang2014gold,ghasemian2015efficient,medina2015sperm}, 
but the sample needs to be in its natural form for motility estimation.
This article focuses on the first step of SQA, image segmentation and detection of non-stained human sperm cells
as analyzed by Ghasemian et al. (2015) \cite{ghasemian2015efficient} and Hidayahtullah et al. (2014) \cite{hidayatullah2014automatic}.
These algorithms apply classical image analysis techniques to solve the problem. To our knowledge no deep learning techniques have been
applied yet.

Our approach focuses on deep convolutional neural networks (CNN) to segment
the sperm cells in the image. There are three main challenges in this approach:
Firstly, every pooling layer in a CNN reduces resolution by at least 50\%; 
after three layers of pooling, every pixel of the result encodes the information of
an 8$\times$8 area of the original image. Secondly, CNNs are often trained
on image patches, however there is a huge class imbalance between background
and sperm pixels, where sperm pixels are significantly harder to detect.
Lastly, we need to cluster the segmentations to objects. Imperfect predictions of the networks
often lead to spurious detections, which need to be removed. One way to do this is to use thresholding
on the size of clusters, leading to an arbitrary threshold parameter.
This parameter needs to be chosen carefully.

We investigate possible solutions to these challenges. While using max-pooling layers
is possible without reducing resolution \cite{giusti2013fast}, an exponential
amount of time in the number of pooling layers is required. This makes it infeasible in practice as the results have to be computed
quickly enough to allow video analysis.
We follow Long et al. (2015) \cite{long2015fully} and investigate up-sampling on the output of the CNN during training and testing. 
Ronneberger et al. (2015) \cite{ronneberger2015u} proposed a more complex architecture, which we disregard since predictions would be too slow
for our application.
Further, we compare training on image patches with training on the full images, where class-labels
are re-weighted to correct the class-skew.
\par For comparison we implemented the sperm head detection method proposed by Ghasemian et al. (2015) \cite{ghasemian2015efficient}.
This method has a similar threshold parameter as our method which has to be adapted for a fair comparison.
For this, we propose a way to adapt the thresholding parameters using
the product of precision and recall on the final detections.

The paper is organized as follows: Section \ref{sec:method} describes the dataset and the CNN architectures used.
Experiments are described in Section \ref{sec:experiments}. Results are given in Section \ref{sec:results} and discussed in Section \ref{sec:discussion}. Finally,
we conclude in Section \ref{sec:conclusion}.

\section{Method}
\label{sec:method}
\begin{figure*}[t]
	\centering
	\includegraphics[width=0.8\textwidth]{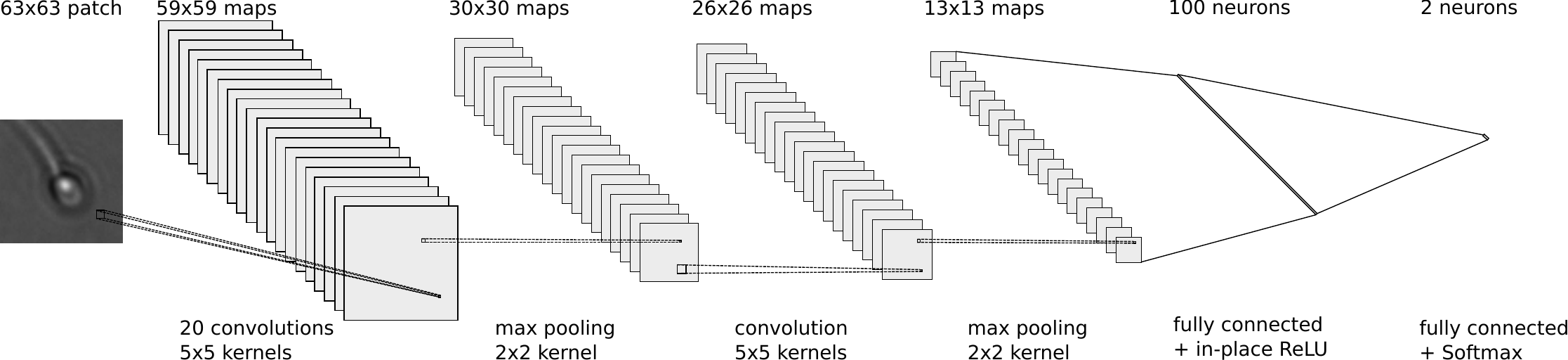}
	\caption{Illustration of the 2-conv CNN}
	\label{fig:2conv}
\end{figure*}
\textbf{Dataset.} We have constructed a dataset of 765 grayscale images of 35 independent sperm samples.
The 35 samples were individually diluted using a solution of Bicarbonate-Formalin (as devised by WHO
\cite{world2010laboratory}) to get an appropriate amount of cells in each image (between 
2 and 290 sperm cells) and to fixate them. Fixation facilitates sedimentation of the cells to the bottom of
the counting chamber, ensuring that all cells are roughly in the same focal plane.
In order to have cells both in and out of focus, reflecting the optical variation, Z-stacks of images were acquired.
The images were acquired using an image cytometer with 20$\times$ optical magnification
and a resolution of 1920$\times$1440 pixels (0.2 $\mu m$/pixel). The image intensities
have been quantized from 14- to 8-bit images. In each image the intensities
where normalized to lie between zero and one.

The images were annotated by experts and registered into two classes: background
and sperm cells. Round cells form an important part of the background and were therefore
also annotated. The tip of the head and the neck point was registered for each sperm
cell while the circumference was annotated for each round cell.
Pixel-segmentation ground truths are generated by creating an
ellipse at the center of each sperm cell head with radius $r_1 = \frac{1}{4} l_{cell}$ and
$r_2 = \frac{2}{3}r_1$ where $l_{cell}$ is the length of the cell head.

We split the samples into 70\% train and 30\% test data
based on stratified sampling on the average number 
of sperm cells in the full images of each sample.
This ensures that images from the same sample are part of the same split as they contain correlated data.
Hence, one sample being part of testing data is never represented in the training data.

From the training dataset we generated an additional dataset of extracted patches from the images using
the annotated classes. This patch dataset contains $63\times63$ pixel patches which are labelled
by their ground truth in the center pixel.
The size of the patches is chosen to allow the entire head, which is typically 25 pixels long, and a small part
of the tail to be included.
From each image, we extract up to 3,000 patches, split into
40\% sperm cells, 40\% background and 20\% round cells. The numbers were chosen
to cover the variety of debris in the background class (round cells contribute a lot to the variability of the background).
Random rotation and flipping is applied before extracting each patch. Table \ref{tbl:data_statistics}
shows statistics for the resulting datasets. Note that the dataset contains a total of 38,708 sperm cells
of which 23,997 are included in the train set and 14,711 are included in the test set.

\begin{table}[tb]
\centering
\begin{tabular}{l *{5}{r}}
	\toprule
	Statistic & Train	& Test & Total \\
	\midrule
	Images & $540$ & $225$ & $765$\\
Sperm cells & $23,997$ & $14,711$ & $38,708$\\
Patches & $1,424,341$ & $601,290$ & $2,025,631$\\
	\bottomrule
\end{tabular}
\caption{Data statistics}
\label{tbl:data_statistics}
\vspace{1em}
\centering
\begin{tabular}{l *{4}{c}}
	\toprule
	Method & $m_{IU}$ & Threshold & $m_{pred}$ (s) \\
	\midrule
	2-conv & $0.6658$ & $200$ & $0.145$\\
2-conv-full & $0.7080$ & $200$ & $0.143$\\
2-conv-full-up & $0.6805$ & $250$ & $0.143$\\
3-conv & $0.6556$ & $200$ & $0.119$\\
3-conv-full & $0.6497$ & $150$ & $0.119$\\
3-conv-full-up & $0.6661$ & $300$ & $\mathbf{0.116}$\\
3-conv-full-up-inc & $\mathbf{0.7387}$ & $150$ & $0.364$\\
baseline \cite{ghasemian2015efficient} & $0.5679$ & $400$ & -\\

	\bottomrule
\end{tabular}
\caption{Experiment results $m_{IU}$, threshold, and $m_{pred}$ for all eight methods}
\label{tbl:stats_means}
\end{table}

\textbf{Networks.} We define seven networks to test against each other.
The first network is called \emph{2-conv}. It is defined for input patches and illustrated in Fig. \ref{fig:2conv}.
It is a standard CNN with two convolutional, ReLU and max-pooling layers 
followed by two fully connected layers separated by another ReLu layer and including 50\% dropout during training. 
The network \emph{3-conv} is obtained by adding an additional
set of convolution, ReLU, and max-pooling layers.
The networks are defined with receptive fields of size 63$\times$63 using 20 filters in each
of their convolution layers and 100 filters in their fully convolutional layer.

For prediction on the full images, the fully connected layers are substituted with fully
convolutional layers as described by Long et al. (2015) \cite{long2015fully} to allow for faster computation.
As each max-pooling layer divides the spatial resolution
of the output by a factor of 2 in each dimension, we further perform bilinear upscaling of the network output 
probabilities to obain a pixel-wise segmentation. 

To compare whether training on full images is beneficial compared to patch-based training,
we define the architectures \emph{2-conv-full} and \emph{3-conv-full},
which have the same structure as \emph{2-conv} and \emph{3-conv} in the
prediction phase and are trained on full images with the final up-sampling removed.
Finally, the architectures \emph{2-conv-full-up} and \emph{3-conv-full-up} also 
incorporate the bilinear up-sampling into the training process.
The networks trained on full images use a receptive field of size 64$\times$64 and the same 
number of filters\footnote{The difference comes from the fact that it is easier
to define a center-pixel in 63$\times$63 receptive fields}.
We further add a network \emph{3-conv-full-up-inc} with the the same receptive field size
but with 64, 128, and 256 filters in the convolution layers and 1024 filters in the fully convolutional layer.
We omit the network \emph{2-conv-full-up-inc} due to limitations in the framework used.

When testing the networks, we perform post-processing of the full size output probabilities in two 
steps: Firstly, we choose the most probable class as output for each pixel. Secondly, we cluster
pixel-wise segmentation to objects by computing the 8-neighbourhood connected components 
and removing components smaller than a threshold $t$. The value of this threshold is found in section \ref{sec:results}.

\section{Experiments}
\label{sec:experiments}

The 2-conv and 3-conv architectures have been trained on the patch dataset and tested on the full image dataset,
whereas all other networks have been trained and tested on the full image dataset. The outputs of 2-conv-full and
3-conv-full are smaller than the label masks of the full images. We therefore downsample the label masks by
factors 4 and 8 respectively. This is done by taking every 4th or 8th pixel corresponding to the center of the
receptive field of the output.

All networks are trained by optimizing the cross-entropy between the predicted and ground truth label.
To compensate for the class skew in the full images during training we re-weight the
classes according to their distribution. The weight $w_i$ of
class $i$ is defined as $w_i = \frac{1}{n_i \sum_j \frac{1}{n_j}}$ where $n_i$ is the number of pixels
belonging to class $i$. Omitting the re-weighting led to far inferior results classifying everything as background.

The architectures have been trained for 200 epochs using the Adam solver \cite{kingma2015adam} with mini-batches
of 256 patches or 1 full image ($1920 \cdot 1440$ ``samples''). 
For training we chose learning rate $\alpha = 0.001$, moment 1 $\beta_1 = 0.9$, moment 2 $\beta_2 = 0.999$, and $\epsilon = 10^{-8}$.
We implemented the networks using Caffe \cite{jia2014caffe}, and the experiments have been
carried out using a single Titan X GPU.

The baseline method \cite{ghasemian2015efficient} consists of three major steps: Noise reduction, object region detection, and sperm head localization.
The method assumes that all sufficiently large object regions are sperm cells and therefore filters out all object regions smaller than a chosen threshold. This
threshold is crucial for the performance of the algorithm and needs to be chosen carefully.

\par On an object level we are interested in finding each sperm cell. For this purpose we use
the two measures $\text{precision} =\frac{TP}{TP+FP}$ and $\text{recall} =\frac{TP}{TP+FN}$, where $TP$ is the number of true positives, $FP$ is the number of false positives and $FN$ is the number of false negatives. A predicted sperm cell is categorized as $TP$ if it covers more than half the area of a ground truth sperm cell.
Each predicted cell can only count as one positive, i.e. a predicted cell covering more than half the
area of two sperm cells counts as one true positive and one false negative. We evaluate precision and
recall for multiple thresholds on the training data to get a precision-recall (PR) curve for every method.
We choose the threshold value that maximizes the product between precision and recall.

Mean intersection over union (mean IU) $m_{IU}$ is used to quantify the pixel-wise segmentation performance
as described by Long et al. (2015) \cite{long2015fully}:
\begin{equation*}
	m_{IU} = \frac{1}{2} \sum_i \left( \frac{p_{ii}}{\sum_j (p_{ij} + p_{ji}) - p_{ii}} \right)
\end{equation*}
Where $p_{ij}$ is the number of occurences of class $i$ predicted as class $j$. We have chosen this measure since it
is invariant to the aforementioned class skew.

\par Finally, fast computations is one of the requirements for automatic SQA. We therefore record the execution
time of computing a prediction and object removal on all 765 full images and compute the mean execution time
$m_{pred}$ per image.
Our baseline method implementation is not as optimized as our networks and therefore we omit the results.

\begin{figure}[H]
\centering
\begin{subfigure}{0.7\textwidth}
\hspace{-0.5cm}
\includegraphics[width=\textwidth]{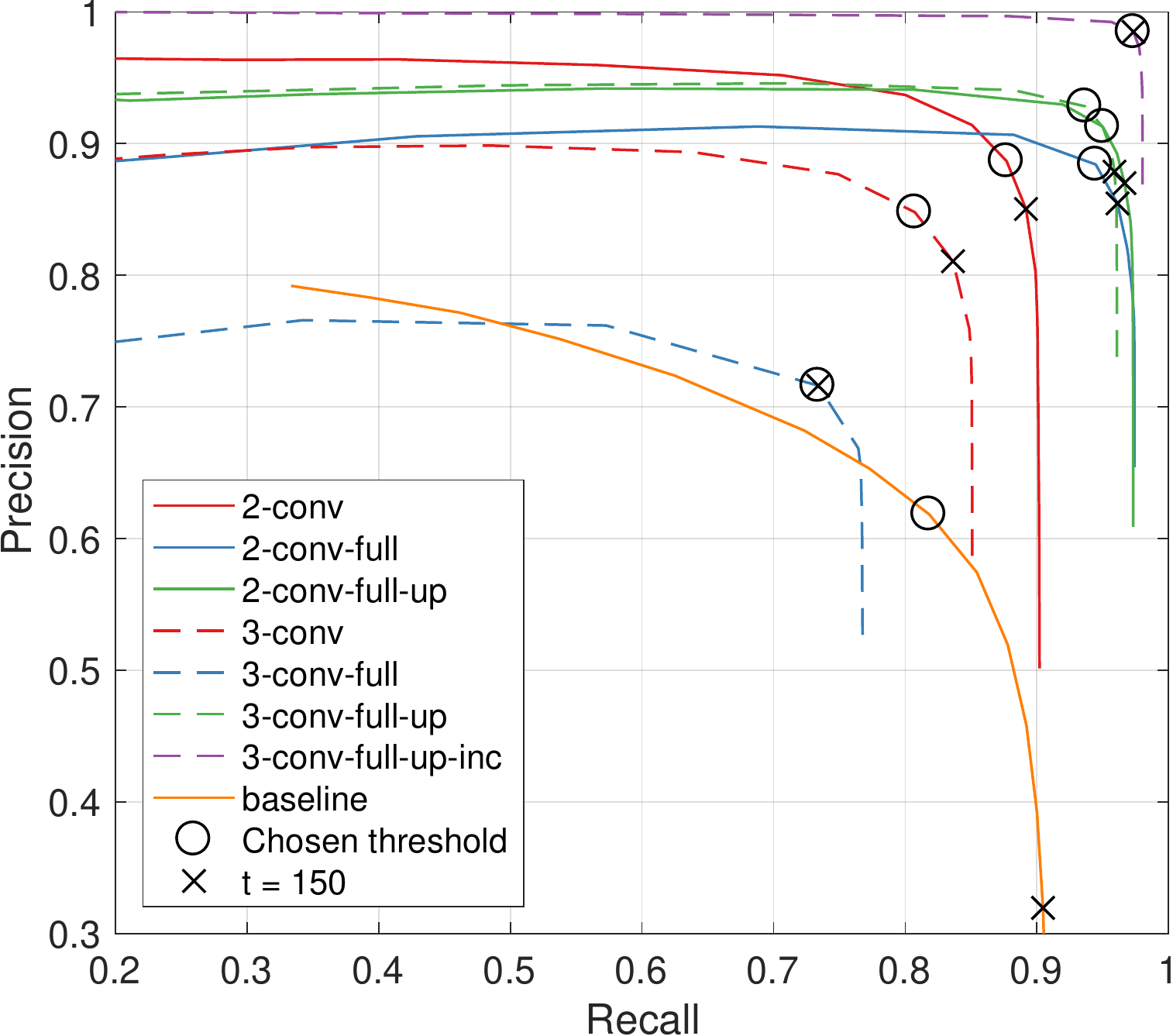}
\caption{}
\label{fig:pr-curve-train}
\end{subfigure}
\begin{subfigure}{0.7\textwidth}
\hspace{-0.5cm}
\includegraphics[width=\textwidth]{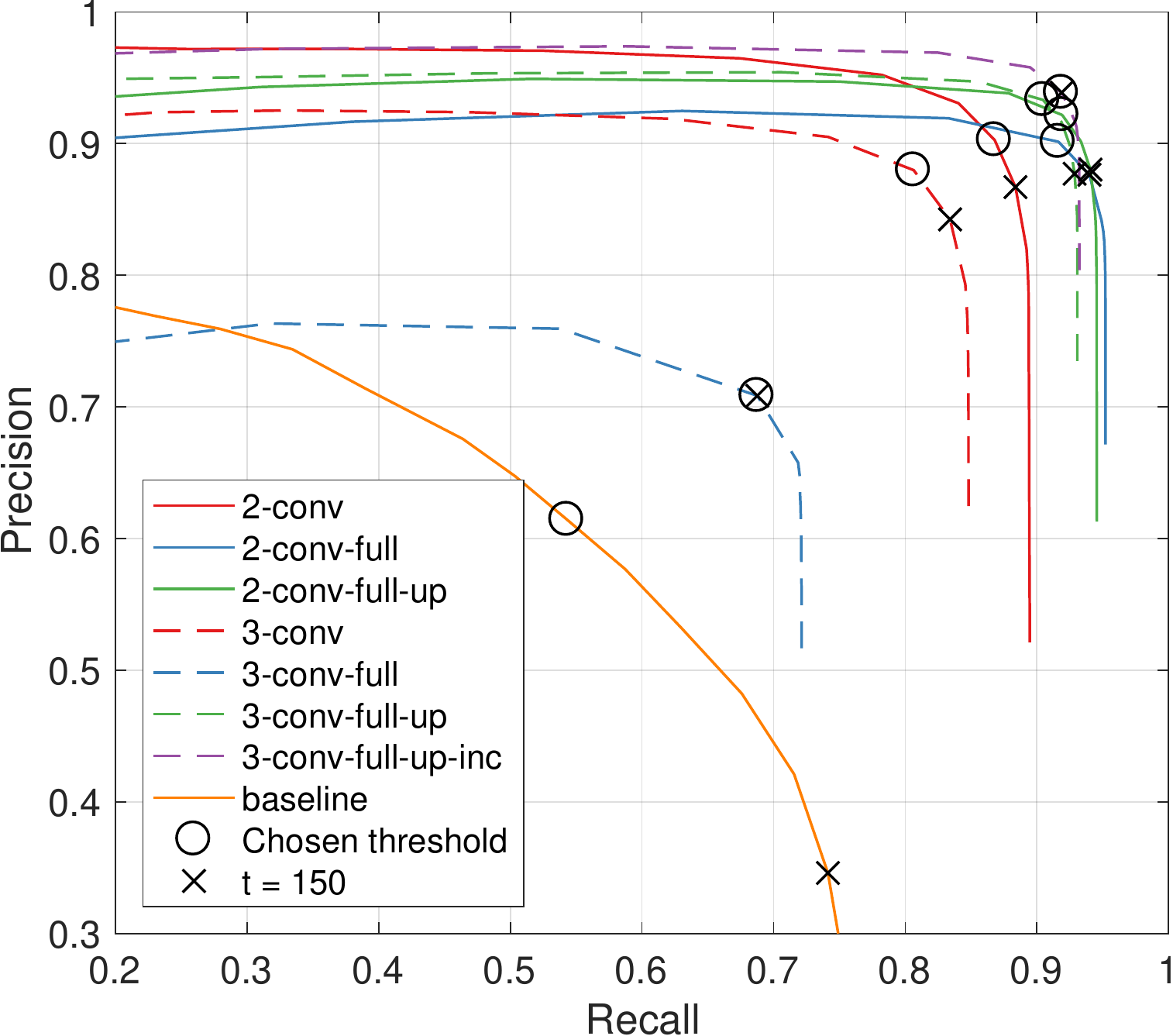}
\caption{}
\label{fig:pr-curve-test}
\end{subfigure}
\caption{Precision-recall graphs for (\subref{fig:pr-curve-train}) train and (\subref{fig:pr-curve-test}) test for all
networks. 2-conv networks are plotted using fully-drawn lines and 3-conv networks using dashed lines.
Same colours of lines indicate same parameters. The black circles indicate the point on
each graph that corresponds to the threshold $t$ reported in Table \ref{tbl:stats_means}, while the black crosses indicate the points for $t = 150$.}
\label{fig:pr-curves}
\end{figure}

\section{Results}
\label{sec:results}
Results of the mean IU $m_{IU}$, thresholds found by maximizing the product between
precision and recall on training data, and mean execution time $m_{pred}$ for each method are given in Table \ref{tbl:stats_means}.
\par Generally when considering $m_{IU}$, the networks trained on full images with up-sampling
perform better than the networks trained on patches. All networks perform better than the baseline method. Training on full images without up-sampling leads to better results for 2-conv-full but worse for 3-conv-full. The 2-conv networks perform better than their 3-conv equivalents. The network \emph{3-conv-full-up-inc} performs best, but it also has a considerably
higher execution time $m_{pred}=0.364$ than the other methods spanning the range of $0.116 - 0.145$ seconds per image.

\par The results for the object detection are given in Fig. \ref{fig:pr-curves}.
The figure shows the precision-recall graphs for (\subref{fig:pr-curve-train}) train and (\subref{fig:pr-curve-test}) test for all methods.
The graphs have ends due to the smallest and largest thresholds considered (0-1,000).
We plot 2-conv networks using fully-drawn lines and 3-conv networks using dashed lines.
Same colours of lines indicate same parameters. The black circles indicate the point on
each graph that corresponds to the threshold reported in Table \ref{tbl:stats_means}, while the black crosses indicate the points for a threshold of 150.
The baseline performs considerably different on the train and test set even though there is no training involved apart from the choice of threshold.
It performs considerably worse than our networks except 3-conv-full.
The best method is 3-conv-full-up-inc having 93.87\% precision and 91.89\% recall on the test set using threshold 150.
While some overfitting can be seen between training and test, it still outperforms the other methods.

\section{Discussion}
\label{sec:discussion}
Our results show that using neural networks is beneficial compared to the classical approach. The large difference
in baseline performance indicate that there is a large variation between samples. We believe that we have captured the
variation of a sperm cell in our train and test sets, but we have not captured all possible combinations of
cells in an entire image. Given our limited number of individual samples, there are some cell concentration differences.
The baseline performance difference is likely caused by these cell concentration differences.
Our networks are not affected by these differences except to the degree expected from overfitting.
All networks except 3-conv-full-up-inc perform almost the same
on train and test data whereas 3-conv-full-up-inc is showing clear signs of overfitting.
This indicates that our networks are
sufficiently complex to cover the variation of the data and that even larger networks are unlikely to generalize better.
As we have not used the test set for model selection, we can expect the performance on the test set to be close
to the true performance.

Up-sampling has different effects on mean IU and object detection. For mean IU detecting object boundaries is important.
As up-sampling is equivalent to blurring it is not beneficial for mean IU when the model is already able to accurately describe the shape of the objects.
This can be seen in the difference in its effect on networks with two and three max-pooling layers. We hypothesize that training using up-sampling gives us true predictions with cluster areas closer to the true size of sperm cells.
This makes it easier to distinguish sperm cells from a specific type of debris (Fig. \ref{fig:sperm_and_debris} \& \subref{fig:deform_debris}) easily
mistaken for the head of a sperm cell but having a slightly smaller area.

\par When omitting up-sampling, there is no general tendency when comparing patch-based and full-image training.
For 2-conv networks, full-image training seems to profit from the increased variation in the data while patch-based
training profits from the weighting of round cells in the background. This can be seen by the differences in precision and recall for
the two methods in Fig. \ref{fig:pr-curves}.

When we compare the PR-curves, we see that the choice of a fixed threshold can be misleading. It turns out that the ranking of
the networks can change depending on the choice of it. However, the chosen thresholds on the training set lead to consistent rankings
on the test set in our case. Introducing the threshold and optimizing it leads to far superior results for all networks compared
to choosing an arbitrary value.
The obtained precision and recall seems reasonable for the purpose of identifying sperm cells in a semen sample, however it needs
clinical testing for verification of its performance in practice.

\section{Conclusion}
\label{sec:conclusion}
In this paper, we have used deep convolutional neural networks for the task of sperm cell segmentation and object detection.
In this task, we are constrained by the computation time as well as the accuracy demands, which
make it harder to train networks with many pooling layers. To mitigate both problems
we explored the use of full image training and up-sampling of the network outputs in order to increase performance.
We specifically investigated thresholding on the size of detected components. Choosing the product of precision and recall
leads to a robust estimate of threshold parameter.
For deeper networks, up-sampling appears necessary to achieve good segmentation and object detection performance.
The same does not necessarily hold for more shallow networks.
\par Our method outperformed a classical image analysis method
which can be considered state-of-the-art.
Overall the system sensitivity and precision are sufficiently high to be valuable for human sperm analysis systems.

\section{Acknowledgements}
This work is partly funded by the Innovation Fund Denmark (IFD) under File No. 4135-00169B. We would like to thank Department of Growth and Reproduction, Rigshospitalet, Denmark, for helping with annotation of our data.

%
%

\end{document}